\documentclass[letterpaper]{article} % DO NOT CHANGE THIS
\usepackage{aaai24}
\usepackage{times}  % DO NOT CHANGE THIS
\usepackage{helvet}  % DO NOT CHANGE THIS
\usepackage{courier}  % DO NOT CHANGE THIS
\usepackage[hyphens]{url}  % DO NOT CHANGE THIS
\usepackage{graphicx} % DO NOT CHANGE THIS
\usepackage{natbib}  % DO NOT CHANGE THIS AND DO NOT ADD ANY OPTIONS TO IT
\usepackage{caption} % DO NOT CHANGE THIS AND DO NOT ADD ANY OPTIONS TO IT

\usepackage{algorithm}
\usepackage{algorithmic}
\usepackage{amsmath}
\usepackage{pifont}
\usepackage{array}
\usepackage{newfloat}
\usepackage{listings}
\DeclareCaptionStyle{ruled}{labelfont=normalfont,labelsep=colon,strut=off} % DO NOT CHANGE THIS
\floatstyle{ruled}
\newfloat{listing}{tb}{lst}{}
\floatname{listing}{Listing}
\newcommand{\cmark}{\ding{51}}
\newcommand{\xmark}{\ding{55}}

\title{Automate or Assist? The Role of Computational Models in Identifying Gendered Discourse in US Capital Trial Transcripts}
\author {
    Andrea W Wen-Yi,\hspace{3mm}
    Kathryn Adamson,\hspace{3mm}
    Nathalie Greenfield,\hspace{3mm}
    Rachel Goldberg,\hspace{3mm}\\
    Sandra Babcock,\hspace{3mm}
    David Mimno,\hspace{3mm}
    Allison Koenecke
}
\affiliations {Cornell University\\
    \texttt{\{andreawwenyi@infosci.,ka468,nmg76,rtg67,slb348,mimno,koenecke\}@cornell.edu}
}
\begin{document}

\maketitle

\begin{abstract}
The language used by US courtroom actors in criminal trials has long been studied for biases. However, systematic studies for bias in high-stakes court trials have been difficult, due to the nuanced nature of bias and the legal expertise required.
Large language models offer the possibility to automate annotation. But validating the computational approach requires both an understanding of how automated methods fit in existing annotation workflows and what they really offer.
We present a case study of adding a computational model to a complex and high-stakes problem: identifying gender-biased language in US capital trials for women defendants.
Our team of experienced death-penalty lawyers and NLP technologists pursue a three-phase study: first annotating manually, then training and evaluating computational models, and finally comparing expert annotations to model predictions.
Unlike many typical NLP tasks, annotating for gender bias in months-long capital trials is complicated, with many individual judgment calls.
Contrary to standard arguments for automation that are based on efficiency and scalability, legal experts find the computational models most useful in providing opportunities to reflect on their own bias in annotation and to build consensus on annotation rules.
This experience suggests that seeking to replace experts with computational models for complex annotation is both unrealistic and undesirable. Rather, computational models offer valuable opportunities to assist the legal experts in annotation-based studies.

\end{abstract}

\maketitle

\section{Introduction}
Researchers studying legal language have long suggested that gender bias plays a significant role in high-stakes court cases. 
But building evidence has been challenging because of the subtle, contingent nature of biases and the vast quantity of legal transcripts.
New language technologies offer the potential to identify and measure biased language at scale.
In this work, we present a case study involving a collaboration between technologists and legal scholars.
Our work highlights both the promise and challenges of using powerful new language models to annotate complex concepts.
In particular, we focus on the interaction between AI systems and human experts, which leads to both a more developed understanding of gender bias in court transcripts and a more effective computational tool for legal researchers.

This paper studies trial transcripts for women defendants facing capital charges in US courts.
The study involves three phases. 
In the first phase, the legal experts identify and annotate the following four themes of gender-biased language (i.e., language that invokes gender stereotypes) regarding the defendant. The themes are developed upon review of the trial transcripts through Critical Discourse Analysis (CDA):
\begin{itemize}
    \item describing the defendant's inappropriate expression of emotions, such as joy or lack of grief
    \item hypersexualizing the defendant
    \item describing the defendant as evil, deceitful or manipulative
    \item characterizing the defendant as a bad mother
\end{itemize}

In the second phase, the NLP experts develop and evaluate computational models to identify the four themes in trial transcripts, following --- for this phase --- the standard NLP practice of assuming that expert annotations are a ``gold-standard.'' 

In the third phase, we seek a synthesis between expert and model annotations. Three legal experts, who have been closely involved with capital trials, re-read passages with large disagreements between expert and model annotations. They discuss whether to relabel those passages and provide reasons for their decisions. Through participating in the decision-making processes of the legal experts, the full team gathers insights about the desired role of computational tools in the eyes of legal experts in this setting. 

The annotation process is time-consuming and difficult --- identifying stereotyped language in capital trials requires knowledge about capital trial process, legal practicality and terminology, defendant demographics, tone of speech, as well as deep familiarity of each case. 
The complexity of the task requires frequent discussions among legal experts to negotiate definitions of themes. This is a task that requires frequent consensus building and cannot be done individually.
In standard NLP protocols, labels are intended to be clear and unambiguous. Identifying and defining biased language is, however, a nuanced task that requires a large degree of subjectivity and interpretation.
But to carry rhetorical weight, evidence for gender bias in trials must be collected and assessed with as much care to avoid personal bias as possible.

In this work, we ask the question: what is the role of computational models, including large language models (LLMs), in this highly laborious, complex, and nuanced annotation process? While using a LLM to replicate expert annotations may seem intuitive, we find that completely replacing human judgment is both unrealistic, due to the frequent inconsistencies in annotations, and undesirable, as experts have no interest in ceding ultimate responsibility.
Rather, experts want both efficiency gains and the opportunity to be challenged to refine and strengthen their argument.
Our concern, therefore, is to study the benefits offered by a finetuned language model in \textit{assisting} the annotation process, as well as its limitations. 

The paper is organized as follows: Section \ref{sec:related-work} outlines the related works in the empirical and theoretical studies of gender and sentencing, and computational approaches in studying gender stereotypes. Section \ref{sec:data} describes transcript collection. Section \ref{sec:quali} presents the method and results of the first phase: qualitative annotation. Section \ref{sec:quanti} discusses the second phase: quantitative modeling. Section \ref{sec:expert-eval} discusses the third phase, where legal experts and technologists revisit passages where model disagrees with annotations from first phase. We discuss our findings in Section \ref{sec:discussion}. 

Through this process, we identify the challenges of using LLMs for the purpose of quantifying gender bias in large, complicated corpora --- we note that there are many nuances that go into replicating the experienced annotations. Furthermore, we derive insights about the benefits of employing imperfect computational models in the process of annotating complex concepts --- the legal experts report that they are not looking for models that perfectly replace the need for them to annotate; rather, they look for new perspective that could challenge their own definitions and remind them of their potential blind spots developed over hundreds of hours of annotation. Lastly, we provide design suggestions for the direction of computational model improvements that fit with the interaction patterns desired by the legal experts. 

\section{Related Work} \label{sec:related-work}

\paragraph{Gender and sentencing.}
Gender disparity in sentencing outcomes is widely studied in the United States. The vast majority of empirical works suggest that women received more favorable sentencing decisions than men in comparable crimes \cite{crew1991sexdiff, embry2012SexOffenders, koerner2014federal, daly1995SexEffects, FarnworthTeske1995GenderFelony}. Scholars have also suggested that such disparities in sentencing vary across crimes; for example, the gender effect is inconclusive in more violent crimes  \cite{RodriguezCurryLee2006GenderDiff, NagelHagan1983GenderAndCrime, koons-witt2014SC}. 

Legal and criminal scholars have developed two major theories for the observed gender disparity in sentencing decisions --- the \textit{chivalry theory} and \textit{evil woman theory}, both of which are based on the patriarchal point of view on binary gender roles. 
The \textit{chivalry theory} suggests that women receive preferential treatment in criminal trials because they are stereotyped as a weaker sex, and hence need to be protected by men from suffering \cite{visher1983chivalry, RodriguezCurryLee2006GenderDiff}. On the other hand, the \textit{evil woman theory} posits that the preferential treatment does not extend to women who violate traditional gender roles, and that women could be punished more harshly than male offenders when committing violent crimes or crimes that are perceived as masculine \cite{rapaport1990some}.

\paragraph{Narrative analysis in legal decision making.}
While there are many studies on the relationships between offender gender and sentencing decisions, the majority of these studies are statistical in nature \cite{streib1990death, koons-witt2014SC, koerner2014federal, RodriguezCurryLee2006GenderDiff}. Researchers have looked into narratives and language used in legal decision-making  such as asylum claims \cite{shiff2021sociology}, parole hearings \cite{GreeneDalke2021You} and rehabilitation treatments \cite{Wyse2013Rehabilitating}. However, these studies on language are mostly derived from close readings of a small number of cases \cite{GreeneDalke2021You, mogul2005-DykierButcherBetter, sutton2022-DeathDehumanizationProsecutorial, dalke2023prerequisites}. Studies on courtroom language in capital trials of women are also rare and on only a handful of women \cite{mogul2005-DykierButcherBetter, sutton2022-DeathDehumanizationProsecutorial}. In this study, we explore the role of computational models in expanding the valuable annotations for systematic investigations of narratives in the courtrooms.

\paragraph{Gendered discourse in US capital trials.}
Nationwide, one of every fifteen women in prison is serving a sentence of life, virtual life, or death \cite{nellis2021extreme}.
The use of biased language in criminal trials has been identified as a concern, especially in race-based stereotyping \cite{baldus1983comparative, 2016-BenettStirling}. Murder convictions and death sentences have been overturned due to the use of racial stereotyping language in criminal trials, highlighting the importance of identifying stereotyping narratives in capital trials \cite{levenson2022Judge, duane_buck_nytimes}. 
Gendered discourse in US capital trials has been less studied and only in the cases of a handful of women on death row \cite{rapaport1991-DeathPenaltyGender, atwell2007wretched}. Yet the stakes are high --- a death-sentenced defendant is actively seeking appeal in the Supreme Court due to courtroom languages based on gender stereotypes \cite{brenda_andrew_nytimes}. 
Most previous linguistic and legal work on the gendered features of courtroom speech is in non-capital criminal trials \cite{potts2018-MotherMonsterMrsa}, civil trials \cite{conley2019just}, or cases where the woman was the victim rather than the defendant \cite{ehrlich2019well, ehrlich2021intersections, potts2021-WomenVictimsMen}. Most research relating to women and capital punishment has focused on the types of crimes for which women have been sentenced to death \cite{baker2015women, streib2002-GenderingDeathPenalty, streib2005-RareInconsistentDeath, carroll1996images, rapaport1991-DeathPenaltyGender} rather than courtroom language.
In this study, we seek to analyze the gendered courtroom language deployed in capital trials of women.

\paragraph{Computational approaches in detecting gender stereotypes.}
Detecting language with sexism and gendered stereotypes using Natural Language Processing (NLP) techniques has drawn a large degree of scholarly interest in recent years \cite{basile-etal-2019-semeval}, with the majority of applications in online content such as social media \cite{mozafari2020hate, Anzovino2018AutomaticIA, Chiril2021BeNT, chiril-etal-2020-said, vasquez-etal-2022-heterocorpus} and news articles \cite{devinney-etal-2020-semi}. 
Unlike trial transcripts, these online texts are very short, usually no more than a few hundred characters, and are meant to be consumed by the general public rather than trained specialists. 

Automated analysis of legal language using NLP has also attracted widespread attention in recent years \cite{Bommarito2018LexNLPNL, chalkidis-etal-2022-lexglue}, with a particular interest in predicting legal outcomes \cite{medvedeva-mcbride-2023-legal, vaudaux-etal-2023-pretrained, barale-etal-2023-asylex, chalkidis-etal-2019-neural}. Scholars have identified that the extremely long sequence length of legal documents, and the legal expertise required to understand the complex terminologies, make automatic processing of legal documents a unique challenge for even the state-of-the-art prompting strategies and LLMs such as GPT-3 and Llama \cite{thalken-etal-2023-modeling, hakimi-parizi-etal-2023-comparative}. Our work takes as given these LLM weaknesses, and focuses on how computational models can still benefit the legal community.

\section{Data description and collection}  \label{sec:data}
As of January 2024, there are 47 cis-gender women on US death row. We obtained trial transcripts for all of them. Building this corpus required personal networking with defense attorneys: while transcripts are technically public, acquiring them can cost thousands of dollars. Most of the transcripts are in PDF format. We preprocess court transcripts with the following steps: first, we use Adobe optical character recognition (OCR) to transform PDF transcripts into text files. Second, we lower case and remove all digits in the text files. Lastly, we use Python package \verb|NLTK| \cite{bird2009natural} to split transcripts into sentences. We only keep sentences with at least three words. While we have $47$ trial transcripts, we focus on $8$ transcripts with the highest OCR quality in this study.

In the United States, there are two phases in death penalty trials: a guilt phase, where the jury determines whether the defendant is guilty of the crime charged, and a penalty phase, where the jury (or, on rare occasions, a judge or panel of judges) decides what sentence is appropriate. At each of these two stages of the trial, the prosecution and defense can make opening and closing statements to the jury, present witnesses, and cross-examine the witnesses who testify on behalf of the other side. We study the entirety of both stages for the eight trials.

While the absolute number of trials we have may seem small, the total size of the collection is massive.
The eight transcripts, documenting months of conversations in court, are each 1,126--3,415 pages long. After preprocessing, the lengths of the eight transcripts range from 178,790 words (roughly the length of Frank Herbert's \textit{Dune}) to 609,774 words, with an average of 351,000 words per transcript.

\section{Qualitative annotation} \label{sec:quali}
In this first phase, legal experts identify gender-biased themes following standard procedures from critical discourse analysis (CDA) \cite{blommaert2000critical}. Broadly speaking, CDA recognizes that discourse both creates and reflects cultural norms. To discern the ways that discourse creates social reality, analysis should ``examin[e] the linguistic practices through which it is translated into social action'' \cite{conley2019just}.

Potentially gendered themes of interest are initially developed by a close reading of the prosecution’s opening statement through the guilt phase by two legal practitioners who have been closely involved with capital trials. These initial themes formed the annotation training for a group of graduate assistants (GA) with legal backgrounds. Then, the GA team annotates $8$ transcripts in their entirety (including opening statements, witness testimony, and both the prosecution and defense attorney's closing arguments in both phases of trial) The team uses the qualitative data analysis software, MaxQDA, for annotation. 

Each transcript is annotated by one GA with a legal background. While there is no inter-rater reliability to report, the annotation team met weekly to discuss and resolve questions that arose over the annotation process. 

The resulting four themes that we explore in this paper are described in Table \ref{tab:theme_subcodes}. The annotated sentences are rare in relation to the entire transcript. Furthermore, each sentence could belong to multiple themes.

Qualitative annotation is extremely time-intensive and expensive. In total, it took the GAs more than $300$ hours to annotate $8$ transcripts, not including the time for training and weekly discussion. Hence, it is infeasible to extend the annotations to all women's transcripts by manual annotation exclusively. In addition to the investment it required, manual annotations have other challenges. For example, we observe instances that were not coded but should be (see examples in Sec \ref{sec:expert-eval}). There are also inconsistencies in the unit of annotations --- while some GAs assign the theme to sentences, others assign it to an entire conversation, within which not every sentence is meaningful (for example, the short responses such as ``yes'' and ``no'' from the witnesses in cross-examinations). 

These observations present opportunities and challenges for computational modeling. Hence, our concern is not building a model to replace the manual annotations. In fact, this is not possible because the input data are contestable from a quantitative point of view.  Rather, we focus on using a computational model to assist the annotation process, resulting in more consistent annotations. 

\begin{table*}
\centering
\begin{tabular}{|llm{7cm}ll|}
\hline
  Theme Name &  Code & Descriptions & \# sentences & mean \% sentences\\
\hline
\hline
Emotions & EMOT & Describing the defendant's emotions, including emotional, emotionless, inappropriate joy or happiness, remorseful, or remorseless. & 255 & 0.136\\
\hline
Hypersexualization & SEX & Describing the defendant as promiscuous, loose, an adulterer; attacking the defendant's sexual expression, or demonizing their sexual practices. & 339 & 0.174\\
\hline
Betrayal of Gender Norms & NORM & Describing the defendant as manipulative, greedy, evil, deceitful. &793 & 0.337\\
\hline
Bad Mother & MOM & Describing the defendant as a bad mother. & 148 & 0.064\\
\hline
\end{tabular}
  \caption{Four themes explored in this paper. Mean \% sentences is the average \% of annotated sentences across the eight transcripts. On average, each theme represents fewer than 1\% of sentences in a trial transcript. }
  \label{tab:theme_subcodes}
\end{table*}

\section{Quantitative modeling} \label{sec:quanti}

In the second phase of the study, we develop and evaluate computational methods to annotate gender-biased language.
This phase follows standard methodology from NLP, treating human annotations as ``gold standard'' data for training and evaluation, despite the underlying data concerns outlined above.\footnote{We release code and appendix at: \url{https://github.com/andreawwenyi/automate-or-assist}}

Since each sentence can belong to more than one theme, we treat each theme as a separate binary classification task. For each theme, we finetune a \texttt{LEGAL-BERT} classifier \cite{chalkidis-etal-2020-legal}. Our choice of model is informed by the current literature in the space of legal NLP. Scholars have experimented with comprehensive prompting strategies with state-of-the-art LLMs on different kinds of legal tasks. Recent works have found that prompted LLMs perform poorly on various kinds of legal tasks \cite{hakimi-parizi-etal-2023-comparative} and that in-domain \texttt{LEGAL-BERT}, albeit smaller, outperforms other larger, newer LLMs on a complex annotation task in the legal space \cite{thalken-etal-2023-modeling}. 
\paragraph{Including context in input and output}
The legal experts seldom only rely on a single sentence to determine if there is a gendered strategy at play. The annotation is often judged in context. For example, the question \textit{``What was the defendant's demeanor like at that time?''} could be related to \textit{Emotions} or not based on the witness's answer or the other questions that come before or after (see Table \ref{tab:emotions-examples}). Therefore, we'd like the model to learn from context when making predictions, and also be flexible in the output length, adjusting based on the content.

As the language model's context length is limited, it is infeasible to feed the entire transcript to a model. We thus have to break the transcripts into manageable chunks of text. We use a sliding window approach with window length of $10$ and a step of $1$. Specifically, an entire transcript is broken down to paragraphs that each contain 10 consecutive sentences and differ from the next paragraph by 1 sentence. This way, each sentence is included in 10 distinct paragraphs.  

We predict whether \textit{at least one sentence in each paragraph belongs to a theme of interest} using binary classification. We calculate the score for each sentence by averaging the scores assigned to the 10 paragraphs containing that sentence.

\paragraph{Finetuning and cross-validation}
Because negative labels are much more frequent than positive labels, we under-select negative paragraphs to be three times the number of positive paragraphs. We apply leave-one-out cross-validation to the $8$ annotated court transcripts. Specifically, we finetune a model on $7$ transcripts, then we apply the trained model on the held-out transcript. This strategy allows us to evaluate how well the model generalizes to unseen transcripts. We finetune \texttt{LEGAL-BERT} hosted on Huggingface with a $2\times 10^{-5}$ learning rate and $0.01$ weight decay.

We also experimented with zero-shot prompting \texttt{FLAN-T5-Large} and \texttt{GPT-3.5}. See details in the appendix.

\begin{table*}    
\centering
\begin{tabular}{|lp{6cm}lp{7cm}|}
\hline
  &Paragraph & EMOT Annotation & Reason\\
\hline
\hline
A & Prosecutor: What was her demeanor like at that time? \newline Witness: I don't recall. & Negative & A neutral question that does not establish a description of the defendant's emotion. \\
\hline
B &Prosecutor: What was her demeanor like at that time?\newline  Witness: She looked calm. & Positive & An exchange that describes the defendant's emotion. \\
\hline
C &Prosecutor: Did she appear to be grieving? \newline Witness: Possibly. \newline  Prosecutor: What was her demeanor like at that time? \newline Witness: I could not see very clearly. & Borderline / Positive & The ``grieving" question is setting up for an answer regarding the defendant's emotions. This makes the demeanor question more than a neutral question but leads to a specific answer about the defendant's emotions. \\  
\hline
\end{tabular}
  \caption{The same question ``What was her demeanor like at that time?'' could result in an annotation of ``Emotions'' or not, depending on the conversation the question is situated in.}
  \label{tab:emotions-examples}
\end{table*}
\subsubsection{Coreference Resolution}
Identifying a theme is necessary but not sufficient: trials have many participants, and thematically relevant language does not always refer to the \textit{defendant} (rather, it can refer to another person).
For example, the sentence \textit{``You could hear the fear in her voice.''} falls under the theme of ``Emotions.'' However, it is not necessarily portraying the defendant---it could be describing a witness. To filter to instances where target is not the defendant, we additionally utilize coreference resolution.

Coreference resolution is a popular NLP task that identifies mentions in texts that refer to the same entity. We run the LingMess model from Python package \textit{fastcoref} \cite{otmazgin-etal-2022-f} on a 20-sentence passage---the target sentence and its 19 preceding sentences. For each entity mentioned in the target sentence, we examine their coreference clusters. We consider the target sentence to be about the defendant if there is one cluster with direct mentions of the defendant, or if there is one cluster that only contains she/her pronouns.

\subsubsection{Quantitative model output}

After we apply the trained \texttt{LEGAL-BERT} on unseen transcripts, we obtain a score for each sentence. We group together consecutive sentences that score above 0.5 as one output passage. This helps us to be flexible in output length that presents meaningful chunks of information.
Then, we use coreference resolution to identify whether the passages refer to the defendant. Specifically, we run coreference resolution on sentences scored above 0.9, and we only keep passages if they contain at least one sentence scored above 0.9 that mentioned the defendant. 

\subsubsection{Results}
The quantitative model performance is not perfect; however, this is expected given the varying quality of annotations and the vastly different contexts across trials.

One metric we use to evaluate the model performance is passage-level precision, calculated as the proportion of passages with at least one sentence annotated positively by experts among passages predicted positive by the model. The mean passage-level precision is 0.151 for \textit{Emotions}, 0.126 for \textit{Hypersexualizations}, 0.272 for \textit{Betrayal of Gender Norms}, and 0.063 for \textit{Bad Mother}. On average, the top three passages with the highest scores from each transcript have a precision of 0.417 for \textit{Emotions}, 0.333 for \textit{Hypersexualization}, 0.458 for \textit{Betrayal of Gender Norms} and 0.083 for \textit{Bad mother}. 
The other evaluation metric we use is sentence-level recall: the proportion of sentences annotated positively by models among sentences annotated positively by humans. The mean sentence-level recall is 0.484 for \textit{Emotions}, 0.341 for \textit{Hypersexualization}, 0.287 for \textit{Betrayal of Gender Norms} and 0.246 for \textit{Bad mother}. Due to the inconsistencies of annotations, passage-level recall is ill-defined and sentence-level precision is a less-than-reliable measure. See details in the appendix.

\section{Expert evaluations on annotator-model disagreements} \label{sec:expert-eval}

Neither the CDA nor the NLP phases are sufficient on their own in producing large-scale annotations.
Manual annotations are too costly to scale to the full 47 trials, and the influence of individual perspectives raise persistent concerns.
In the third phase, we synthesize results from the previous two phases.
Legal experts find that the contrast between human and machine annotations was useful and prompted good discussion.

For each theme, three lawyers, who have experience representing defendants in capital trials, examine about $30$ passages where the model predictions differ from the graduate assistant (GA) annotations. 
Specifically, the first author selects the three highest-scored passages (meaning that the model finds them to be highly relevant) that the GA annotated as irrelevant from each trial, along with the 6--8 lowest-scored passages that the GA annotated as relevant. 
For each passage, three coauthors who are lawyers jointly decide whether at least one sentence within the passage is relevant to the theme. They select from ``Positive (\cmark),'' ``Negative (\xmark)'', or ``Undecided (\textbf{?})'' (if they are unable to reach a consensus). They also provide reasons for their decisions.
Before the lawyers decide on a passage, they are not told the GA or model's annotation of the passage---they only know that the GA and model annotations disagree. This is to avoid potential biased preferences towards the GA or model's decision. The first author moderated and documented the discussion processes. 

In meetings across three days, the lawyers reviewed passages from the four themes in the following order: \textit{Emotion}, \textit{Hypersexualizations}, \textit{Betrayal of Gender Norms}, and \textit{Bad Mother}. The process is carried out using an annotation tool called Prodigy.
\subsubsection{Results}
\begin{table*}
\centering
\begin{tabular}{|l|l|l|l|m{3cm}|m{3cm}|}
\hline
  Theme & \# minutes & \# passages (\# FP, \# FN) &  \# sentences & \# positive, \# negative, \# undecided in FP &  \# positive, \# negative, \# undecided in FN \\
\hline
\hline
EMOT & 40 & 28 (22, 6) & 359 & \textbf{16}, 5, 1 & 6, \textbf{0}, 0\\ 
SEX & 65 & 30 (24, 6) & 378 & \textbf{0}, 21, 3& 3, \textbf{3}, 0\\
NORM & 35 & 32 (24, 8) & 402 & \textbf{13}, 7, 4& 6, \textbf{1}, 1\\
MOM & 42 & 26 (18, 8) & 479 & \textbf{2}, 12, 4 & 6, \textbf{1}, 1\\
\hline
\end{tabular}
\caption{Quantitative results of the lawyers' evaluations on passages where GA annotator and language model disagree. Numbers where the lawyers agree with the model decision are bolded. FP (False Positives) refers to passages annotated negatively by the GA but positively by the model; FN (False Negatives) refers to passages annotated positively by the GA but negatively by the model.}
\label{tab:expert-quantitative}

\end{table*}

\begin{table*}
\centering
\begin{tabular}{|l||l|l|l|l|}
\hline
& EMOT & SEX & NORM & MOM \\
Reasons for Lawyers' Disagreement with Model Decisions &  &  &  &  \\
\hline
\hline
Unrelated to theme & 1 & 13 & 1 & 7\\  

Related to theme but not describing the defendant. & 3 & 3 & 3 & 4\\  

Neutral information gathering or factual statements of cases & 1 & 5 & 2 & 4\\ 
Need longer context & 1 & 3 (+5$^\star$) & 4 & 0 \\
Defense' counter argument to the theme & 0 & 0 & 1 & 1 \\
\hline
\end{tabular}
\caption{We categorize why lawyers disagree with model decisions into five categories. Usually, they highlight one main reason, so the categories are mutually exclusive. In the SEX theme, there are five cases where \textit{longer context} is cited as a secondary reasons, we marked them with a star ($^\star$) for emphasis.}
\label{tab:expert-disagree-reasons-breakdown}
\end{table*}

We analyze the model-lawyer agreement across four themes. We define model-lawyer agreement in each theme as the number of passages where the lawyers agree with the model predictions, divided by the total number of passages that the lawyers read. Additionally, we analyze the time taken for the lawyers to review passages for each theme. We use \textit{False Positives} (FP) to refer to passages annotated negatively by GA (from the first phase) but positively by the model (from the second phase), and \textit{False Negatives} (FN) to refer to passages annotated positively by GA but negatively by the model. The main metrics are shown in Table \ref{tab:expert-quantitative}.

\subsubsection{Emotions}
Model-lawyer agreement is the highest in \textit{Emotions}, at 57.14\%. Among FP passages, the lawyers agree with 72.7\% of model decisions. This shows that the model has a good understanding of this theme, and that employing the computational model improves the quality of annotations. The lawyers have the least disagreements with each other in \textit{Emotions}, with only $1$ \textit{undecided} passage. Words of emotions, such as \textit{calm, mad, agitated, upset, remorse} are often cited as justifications for decisions, per Table \ref{tab:expert-qualitative-accept}.

However, challenges for computational models arise when one sentence contains references to multiple people. For example, 
\begin{quote}
    \emph{
    [The defendant] said that [the victim] looked really mad, really upset.
    }
\end{quote}
Additionally, when the passage describes the \textit{lack of} emotions, the model often makes false negative predictions. For example, 
\begin{quote}
    \emph{
Q. Do you recall what was said? \\
A. [The defendant] was talking about how she had a lot
of food at the house that had been brought in by friends
and family members and that she should have brought
that to the sheriff’s office because she was sure we were
hungry and would want something to eat. \\
Q. Okay. Anything else you remember she said? \\
A. There was really nothing else, I was so taken back by
that, that after being arrested for the capital murder
of her daughter, that she was worried about food.}
\end{quote}
For more quotes and decisions, see Table \ref{tab:expert-qualitative-emotion} in the appendix.

\subsubsection{Hypersexualization} 
Model-lawyer agreement is lowest in \textit{Hypersexualization}, with lawyers agreeing with only 10\% of model decisions. 
The lawyers also spent the longest time discussing passages of \textit{Hypersexualization}. Compared to the other three themes, it took the lawyers 1.5--2 times longer per sentence.

\textit{Hypersexualization} presents challenges for both the lawyers and the model because of the extensive context required (Table \ref{tab:expert-disagree-reasons-breakdown}). 
The lawyers frequently mention that they need to know information such as the relationships between people in the case and the significance of events being discussed. For example, while reading the following exchange between the prosecutor and a witness, the lawyers mentioned they need to know the relationship between the witness and the defendant, and whether \textit{the brother's wife} referred to the defendant.
\begin{quote}
\emph{
Q: How long were you having sex with your brother's wife?\\
A: From about five years.\\
Q: So you had sex with your brother's wife for five years; how many times?\\
A: Not so much.\\
Q: What does, ``not so much'' mean; 2, 20, 100?
}
\end{quote} The ambiguity of passages led to prolonged discussion time --- the lawyers spent considerable time proposing hypotheses about why a particular topic was discussed in trials. 

The computational model incorrectly picks up passages describing relationships but not necessarily related to sexualization. For example, 
\begin{quote}
    \emph{[The defendant] had a job for a period of time as a greeter at [a supermarket]. This was the end of May, into the beginning of July. So about a little more than a month. So that is where she met [her second husband]. She took up with him, and they were married in December.}
\end{quote}
Additionally, the model also picks up passages that the legal experts deem as neutral information gathering or description of crime facts that are not necessarily a manifestation of gender bias. For example, 
\begin{quote}
    \emph{
   The defendant returned to the residence [and] discovered her ex-husband's body just inside the main door leading from the garage. The defendant called 911, and feigned hysteria. The defendant, in her letters to [her lover], had discussed how she would fake grief upon discovering that her ex-husband had been killed.}
\end{quote}
For more quotes and decisions, see Table \ref{tab:expert-qualitative-hypersexualization} in the appendix.

\subsubsection{Betrayal of Gender Norms}
Overall model-lawyer agreement is at 43.75\%, showing good but slightly lower alignment than \textit{Emotions}. Among the FP passages, the lawyers agree with the model's positive predictions 54.17\% of the time, showing that the model is helpful in picking up passages that the GA missed annotating in the first phase. 

Similar to \textit{Emotions}, direct descriptions of defendants using words such as \textit{greedy, manipulative, evil, calculated} are often cited by the lawyers as annotation justifications of relevance. For example, 
\begin{quote}
    \emph{
    [The defendant], the woman who took every opportunity to line her pockets, a heartless schemer who manipulated and lured men to their peril. She deceived her husband, children, family and friends. She has earned the title premeditated murderer, queen of greed and evil.
    }
\end{quote}

Meanwhile, courtroom actors are likely to describe events or use other indirect descriptions to set up an image of the defendant, resulting in disagreements between lawyers and the need for additional context. For example, while the following exchange between the prosecutor and a witness does not seem relevant in isolation, some lawyers suggest that it could be setting up an image of the defendant being performative.
\begin{quote}
\emph{
Q. Now, was there anything characteristic about [the defendant] that would catch your attention? \\
A. Whenever she would drive her car, she would always have her window down and she would always be smoking her cigarette. And just the mannerism that she would hold her cigarette to the side. \\
Q. How did she do that? \\
A. Just by flicking her cigarette. \\
Q. Any particular style that you refer to it as? \\
A. Well, as I said before, the kids told me it’s like a Hollywood style.
}
\end{quote}
For more quotes and decisions, see Table \ref{tab:expert-qualitative-norms} in the appendix.

\subsubsection{Bad Mother}
Model-lawyer agreement of \textit{Bad Mother} is 11.5\%, similarly low as \textit{Hypersexualization}.
The lawyers struggled to reach consensus with each other: 19.2\% of passages are \textit{undecided}, the highest among the four themes. The passages identified positively by the computational model sparked discussions among the lawyers about decoupling factual statements of crime and value judgements towards the defendants, specifically in trials where the victim was the defendant's child. 
To illustrate, we present two passages that were annotated as positive by the GA annotator but as negative by the model. For the first passage, some lawyers considered it to be \textit{factual} in phase three, leading to be classified as ``undecided'':
\begin{quote}
    \emph{This woman had been trying to get [a man] to kill her daughter for at least months. It got to the point that he went to this child's father and said [the defendant] won't leave me alone about killing [her daughter].}
\end{quote}

For the second passage, lawyers all agreed that it involves a value judgement that a mother should care about her daughter more than herself, leading to a ``positive'' decision in phase three: 
\begin{quote}
    \emph{There has never been a truer statement, by this mom (the defendant). 
That's the way she felt: [My daughter] is ruining my life. She is ruining my life. 
Because this was about [the defendant]. \\
Not about [the defendant's daughter]. \\
This is about [the defendant]. }
\end{quote}  

Some lawyers suggest that it is fair to discuss crime facts in trials. Therefore, while it is intrinsic to the fact that a woman killing her own children is a ``bad mother'', such discussion in trial should not be identified as a manifestation of gender bias. Reading several similar passages, the lawyers went back and forth negotiating differences between factual statements and value judgements.
For more quotes and decisions, see Table \ref{tab:expert-qualitative-mom} in the appendix.

\begin{table*}
\centering
\small
\begin{tabular}{|l|p{7cm}|p{2cm}|p{5cm}|}
\hline
  Theme & Passage & Decisions & Lawyers' reasons \\
\hline
\hline
EMOT & It is beyond comprehension to me that this woman could do what she did, and I submit to you that if you watched her during the course of this trial she has shown no remorse.  & 
GA = \xmark \newline
Model = \cmark \newline
Lawyers = \cmark & 
Describing that the defendant has not shown any remorse. \\
\hline
EMOT & Q. First of all, Mr. [witness], let's talk about the defendant's attitude and emotional state and what was going on when she came back to your house after [the victim] had left with her. She is accusing [the victim] of taking her money. Is she a little bit more agitated at this point? […] She tore your bedroom apart, didn't she? \newline
A. Yes, she did. […] \newline
Q. So, this calm demeanor that [the defendant] usually had was not present? \newline
A. It wasn't there.  & 
GA = \xmark \newline
Model = \cmark  \newline
Lawyers = \cmark  & 
Portraying the defendant as agitated, not calm.\\
\hline
SEX & Q. Do you recall [the defendant] speaking to you about [her husband] in any other negative ways? \newline
A. Mostly, she just complained he didn't make enough money. It was never enough. […] \newline
Q. How often would she complain to you about [her husband] not making enough money?\newline
A. I don't know exactly how often, but more often than not. […] Whenever it just wasn't enough for her, she would complain. & 
GA = \cmark\newline
Model = \xmark\newline
Lawyer = \xmark&
This is describing a greedy woman. It discusses the relationships between the defendant and her husband, but does not sexualize the defendant. \\
\hline
NORM & [The defendant], the woman who took every opportunity to line her pockets, a heartless schemer who manipulated and lured men to their peril. She deceived her husband, children, family and friends. She has earned the title premeditated murderer, queen of greed and evil. & 
GA = \xmark\newline
Model = \cmark \newline
Lawyers = \cmark& 
Portraying the defendant as a person who manipulates, lies, is greedy and evil, etc. \\
\hline
NORM & Q. Did you see any rings? \newline
A. I saw one or two small ones, yes. \newline
Q. What about earrings, i am pointing to my earrings; do you see them? \newline
A. Yes, i do know earrings, yes. \newline
Q. Did you see any earrings? \newline
A. I didn't see earrings. \newline
Q. What else did you see other than these documents, credit cards, the jewelry, three cell phones, and some purses and coins? & 
GA = \cmark\newline
Model = \xmark\newline
Lawyers = \xmark& 
Do not see the relevance of this conversation --- need more context and background knowledge about this case to know why this was originally coded as related to describing the defendant being ``greedy". \\
\hline
MOM & [The defendant] will pose no danger to children if she's punished to life in prison. [The defendant] wanted to be a mom, and she had four kids. 
    Being a mom, that was her dream, but she was never given the necessary tools to do this right. All four of her kids were taken by CPS the same day [the victim] died because the home was so disgusting and unsafe to live in. &
    GA = \xmark\newline
    Model = \cmark\newline
    Lawyers = \cmark
&
    Portraying the defendant being an unfit mother as ``the home was so disgusting and unsafe to live in''.\\
    \hline
    
\end{tabular}
\caption{Example of passages where the lawyers agreed with model predictions, and the reasons why they agreed.}
\label{tab:expert-qualitative-accept}
\end{table*}

\section{Discussion} \label{sec:discussion}
\subsubsection{Why language models struggle to replicate annotations by legal experts}
Identifying gender-biased language in trial transcripts is a hard task even for legal experts. Even engaging in joint discussions, there are still multiple passages where experts could not reach consensus. We find that it is a hard task for language models because legal expertise is required extensively in the annotation process. We share three instances where computational models lack the necessary legal expertise.

First, the model often makes false positive annotations on passages that the lawyers deem as neutral information gathering or factual statements of crime. Distinguishing the differences between factual statements and value judgements is often not an easy task, even for legal experts. 
Second, discerning subtle tonal differences is another instance that requires legal experience. For example, the lawyers mention the sentence \textit{``she now goes and has another baby''} implies value judgement, but \textit{``She has another baby''} does not. Similarly, \textit{``So you didn't see what her demeanor was at that point in time.''} is a neutral question, compared to the prompting version: \textit{``Did you notice if she was particularly emotional at the time?''} 
Third, extensive background knowledge of individual cases is often necessary to comprehend who and what is being discussed, and to determine the importance of the discussion. This is especially pronounced when the theme involves discussions of relationships, such as in \textit{Hypersexualizations}.

\subsubsection{Benefits and challenges of employing computational models in a complex legal annotation pipeline}
There are multiple benefits in employing a computational model in the annotation process. First, it accelerates the annotation process. Trial transcripts are long but the majority of the transcripts are irrelevant. Therefore, using a computational model in the process to filter out the irrelevant parts of the transcripts greatly improves the efficiency of annotation and enables having multiple legal experts read the same passages for inter-rater reliability.
Second, it increases the accuracy of annotation by identifying passages that annotators might overlook. It also helps the legal experts to adjust annotation instructions. For example, upon reviewing the \textit{Emotions} passages, the legal experts found that the GA annotators frequently missed passages describing ``lack of remorse.'' Hence, they decided to add a new subcategory, ``remorseless'', under the \textit{Emotions} code to assist future annotators. 
Third, the false positive predictions that the model makes provide opportunities for the legal experts to refine annotation rules. For example, upon reviewing \textit{Bad Mother}, they identified the need to clarify the difference between factual statements and value judgements. They decided to create a separate coding category to highlight factual descriptions of crimes where the victim was a child of the defendant.  

The legal experts also identified challenges when annotating \emph{with} model-generated annotations. Specifically, they identify the need for more context: \textit{``Machine annotation is done in isolation, (we) lost the context for decision making.''} --- it could be hard for them to judge a passage in isolation, especially for themes that often involve multiple people or rely on event descriptions to build up a story. Another challenge is the lack of flexibility. The legal experts mention that when they are doing qualitative annotations on the whole transcript on \texttt{MaxQDA}, they are able to go back and forth to revise previous annotation decisions when they see a running theme. However, reading isolated passages identified by the computational model on \texttt{Prodigy}, they have to make decisions as-is, based on the passage shown.  These serve as a good pointer for the design of an interactive annotation pipeline. 

\paragraph{Desired role of a computational model}
In this highly complex annotation task, we find that, first, the legal experts desire a computational model that prioritizes \textit{recall} over \textit{precision}. Second, they appreciate a model that surfaces relevant segments rather than perfect predictions. We discuss these two points in detail.

First, expert annotators would rather see more non-relevant results than potentially miss a relevant result. Therefore, it is better to have a model that retrieves the majority of relevant outcomes, even if it means more irrelevant outcomes are also retrieved. In standard machine learning terms, they prioritize \textit{recall} over \textit{precision}. We caution that this preference for recall may be related to our evaluation setup ---
it is easy to check whether segments already picked out by human annotators are relevant, but challenging to determine whether non-annotated segments are truly irrelevant. Based on the characteristics of these potentially flawed annotations, recall is a more reliable measure than precision.

Second, the flaws of manual annotation are precisely where the legal experts could use a computational model surfacing relevant segments. As one lawyer coauthor mentioned in phase three: \textit{``The model adds another layer of perspectives. I see [the disagreement of model from the previous annotator] as a benefit --- it helps remind me that I may have my own biases. Reading these tricky passages helps me identify my blind spots.''} That is to say, the benefit of the computational model lies in the imperfect predictions it makes.  These imperfect predictions prompt the legal experts to refine, concretize, and build consensus for the annotations. And as pointed out by several lawyers, \textit{``[the model output] is a great educational tool when we onboard new coders.''}. 

\section{Conclusion}

Identifying examples of themes by annotating passages is central to social science research and a significant source of effort and cost. Computational methods, like the language model used in this study, are a promising way to increase scalability and improve annotation consistency.
Standard NLP protocols often seek to build systems that \textit{replace} human annotators.
In contrast, this study focuses on legal experts, who neither intend nor desire to be removed from the annotation process.
In the standard NLP protocol, the primary goal is accuracy, and it is assumed that labels are stable, well-defined, and consistently applicable. Identifying gender-biased narratives in courtroom, on the other hand, is a nuanced task involving subjective decision-making. We identify the value of an \textit{imperfect} computational model in promoting the consistency of annotations in this complex setting. 
Rather than seeking to \emph{replace} experts, we suggest using computational models to \emph{assist} experts. One of the main concerns of experts in generating their own annotations is the influence of their personal bias, leading to biased and inconsistent annotations within a multi-annotator team. The benefits of a computational model, in the eyes of these experts, is to provide valuable opportunities for them to reflect on their own biases, and a space --- grounded in specific examples --- for them to engage in discussions with each other to build consensus on the definitions of annotations.

\section*{Ethical Statement}
\paragraph{Ethical considerations} 
We view our work as a dialogue between computational tools and close qualitative readings, where close reading is an essential part of understanding the gendered discourse in capital trials. We recognize that, though court documents are in the public record, sharing the narratives deployed in a trial could negatively impact the subjects studied in this paper as well as other personnel related to the cases. Topics discussed in a capital trial could involve information regarding deeply private information such as family history, childhood experiences, and relationship history about the defendant as well as witnesses. The fact that we share quotes as examples of gendered discourse gives an additional spotlight to the very kind of language we hope to mitigate. To ameliorate this, we anonymize cases by sharing only background information relevant to the analysis (e.g., excluding names and information such as court location.) In doing so, we aim to minimize the potential harm of linking quotes to defendants. Lastly, due to the ethical and privacy concerns of the defendants being studied, we do not release the raw data for this paper, and only release code for reproducibility. 

\paragraph{Positionality} 
The research question in this paper is primarily concerned with the treatment of women in the criminal legal process. The majority of the research team is comprised of women researchers. While our research team does not consist of individuals who have been defendants in criminal trials, the team does consist of legal practitioners who have been closely involved with capital trials for decades. Our team additionally consists of computer scientists who have previously conducted research on fairness topics. 
\bibliography{main}

\appendix

\section{Other models and training strategies}

\paragraph{Zero-shot prompting with instruction-tuned LLMs}
We experimented with zero-shot learning with instruction-tuned LLMs. Due to data privacy considerations, we avoid using closed source \texttt{GPT-3.5} as our computational model. However, to gauge its performance, we created synthetic data by paraphrasing real trial conversations. We prompted the model with a short case summary and a sentence to see if the model could correctly predict whether a given sentence should be annotated as \texttt{Bad mother} (see Table \ref{tab:gpt}). Our experiment results show promise. While the model made mistakes when there is not enough context about who was talking, in general, the model correctly identified whether the sentence given mentioned parenting of the defendant, and whether it is related to the facts of crime. 

Given this finding, we explored the performance of a smaller open-source instruction-tuned model, \texttt{FLAN-T5-large}, that could fulfill our privacy requirements.
We provides a 1--2 sentence trial summary (SUMMARY), a paragraph (CONTEXT), and a target sentence in the paragraph (TARGET). The results are mixed. We find that a short case summary and context aid the model in identifying people in the paragraph, but the model's answer is sensitive to prompts. In general, the model failed to provide reasons and failed when asked to do more than identifying people, such as to judge if the target sentence described the defendant as a bad mother (see Table \ref{tab:flan-t5-large}). 
 
\section{Evaluation metrics for quantitative modeling} 
As mentioned in Section \ref{sec:quali}, there are differences in the unit of annotations --- some associate a label to a paragraph, while others with a sentence. These inconsistencies lead to difficulties in finding a reliable measure of computational models. Sentence-level precision is an unreliable measure, because there are many sentences annotated by humans that are not standalone meaningful. On the other hand, we are unable to define passage-level recall. Recall measures the proportion of true positive instances that are predicted positive by the model. However, since there are cases when a label is associated with a sentence rather than a well-defined multi-sentence passage, we are unable to define what a ``true positive passage'' is. See the number of sentences annotated positive by GAs, and precision measure broken down for each theme and trial, in Table \ref{tab:precisions}.

\begin{table*}
\centering
\begin{tabular}{|p{0.4\linewidth}|p{0.3\linewidth}|p{0.2\linewidth}|}
\hline
Prompt & Model output & Correct? \\
\hline
\hline
Given a case summary of a capital trial, determine if the snippet of an exchange during trial is (1) talking about parenting by some person. If true, please continue to determine (2) if the parent is the defendant or another individual involved in the case." \newline

Summary: The defendant, Jane, is charged for murder of her husband, Tony, by gunshot for insurance policy proceeds.  \newline

Trial Snippet: ``When she is angry, Jane perceived it as her mom standing up for her."
&
Talking about parenting by some person: True
\newline

If the parent is the defendant or another individual involved in the case: Another individual involved in the case
&
Correct
\\
\hline
Given a case summary of a capital trial, determine if the snippet of an exchange during trial is 1) talking about parenting by some person. If true, please continue to determine (2) if the parent is the defendant or another individual involved in the case."\newline

Summary: The defendant, Jane, is charged for murder of her husband, Tony, by gunshot for insurance policy proceeds.  \newline

Trial Snippet: ``I dropped Jane off at school the other day, and when I went back to pick her up, I found that she was quite intimate with her high school sweetheart." &
Talking about parenting by some person: True\newline
The snippet mentions dropping Jane off at school, indicating involvement in Jane's life as a parent or guardian.\newline
If the parent is the defendant or another individual involved in the case: Another individual involved in the case\newline
Since the snippet refers to dropping Jane off at school, the parent mentioned is someone other than the defendant. & 
Incorrect. In the real case where we paraphrased this snippet from, the person who dropped Jane off at school was her partner, not her parent. 
\\
\hline
Given a case summary of a capital trial, determine if the snippet of an exchange during trial is (1) talking about parenting by some person (True, False, Need additional information). If true, please continue to determine (2) if the parent is the defendant or another individual involved in the case. (3) If so, is the parenting related to the facts of crime \newline

Summary: The defendant, Jane, is charged for murder of her colleague, Tony. \newline

Trial Snippet: ``When Jane is home alone with children, she didn't care about them. She just played on her computer all day long. "
&
Talking about parenting by some person: True (Confidence: High) \newline

If the parent is the defendant or another individual involved in the case: The defendant (Jane) (Confidence: High) \newline

If so, is the parenting related to the facts of the crime: Need additional information
&
Correct \\
\hline
Given a case summary of a capital trial, determine if the snippet of an exchange during trial is (1) talking about parenting by some person (True, False, Need additional information). If true, please continue to determine (2) if the parent is the defendant or another individual involved in the case. (3) If so, is the parenting related to the facts of crime. \newline

Summary: The defendant, Jane, is charged for murder of her daughter.\newline

Trial Snippet: ``When Jane is home alone with children, she didn't care about them. She just played on her computer all day long. "
 &
 Talking about parenting by some person: True (Confidence: High)
 \newline
 
If the parent is the defendant or another individual involved in the case: The defendant (Jane) (Confidence: High)
\newline

If so, is the parenting related to the facts of the crime: Likely (Confidence: High)

&
Correct \\
\hline
\end{tabular}
\caption{Experiments from zero-shot prompting \texttt{GPT-3.5}. We omit the explanations that the model provided except for the one with incorrect prediction.}
\label{tab:gpt}
\end{table*}

\begin{table*}
\centering
\begin{tabular}{|p{0.6\linewidth}|l|p{0.2\linewidth}|}
\hline
Prompts & Model Output & Correct? \\
\hline
\hline
SUMMARY: Sophia was convicted of hiring a family friend, Jack, to shoot and kill her own daughter 21-year-old Nancy. \newline

CONTEXT: Nancy is dead. Nancy was murdered by Jack. but Nancy is dead because of one thing and one thing only, without Sophia, i submit to you, ladies and gentlemen, we expect the evidence to show without Sophia, there is no crime. there is no crime. the woman who carried her in her womb for nine months, listened to her heartbeat, her mother, her finger was on that trigger just like Jack's was. \newline

TARGET: Nancy is dead.\newline

Based on SUMMARY and CONTEXT, does the TARGET sentence mention the defendant in this case? Who is it? &
``yes'' & 
Incorrect, and the model did not provide a name even though it answered "yes".\\
\hline
SUMMARY: The defendant in this case is Sophia. Sophia was convicted of hiring a family friend, Jack, to shoot and kill her own daughter 21-year-old Nancy.\newline

CONTEXT: Nancy is dead. Nancy was murdered by Jack. but Nancy is dead because of one thing and one thing only, without Sophia, I submit to you, ladies and gentlemen, we expect the evidence to show without Sophia, there is no crime. There is no crime. The woman who carried her in her womb for nine months, listened to her heartbeat, her mother, her finger was on that trigger just like Jack's was.\newline

TARGET: Nancy is dead.\newline

Based on SUMMARY and CONTEXT, does the TARGET sentence mention the defendant in this case? Who is it? &
``no''&
Correct\\
\hline
SUMMARY: Sophia was convicted of hiring a family friend, Jack, to shoot and kill her own daughter 21-year-old Nancy.\newline

CONTEXT: Nancy is dead. Nancy was murdered by Jack. but Nancy is dead because of one thing and one thing only, without Sophia, I submit to you, ladies and gentlemen, we expect the evidence to show without Sophia, there is no crime. There is no crime. the woman who carried her in her womb for nine months, listened to her heartbeat, her mother, her finger was on that trigger just like Jack's was. \newline

TARGET: The woman who carried her in her womb for nine months, listened to her heartbeat, her mother, her finger was on that trigger just like Jack's was. \newline

Based on SUMMARY and CONTEXT, who is "the woman" in the TARGET? Give a name. & 
``Sophia'' & 
Correct
\\
\hline
SUMMARY: Sophia was convicted of hiring a family friend, Jack, to shoot and kill her own daughter 21-year-old Nancy.\newline

CONTEXT: Nancy is dead. Nancy was murdered by Jack. but Nancy is dead because of one thing and one thing only, without Sophia, I submit to you, ladies and gentlemen, we expect the evidence to show without Sophia, there is no crime. There is no crime. The woman who carried her in her womb for nine months, listened to her heartbeat, her mother, her finger was on that trigger just like Jack's was.\newline

TARGET: The woman who carried her in her womb for nine months, listened to her heartbeat, her mother, her finger was on that trigger just like Jack's was.\newline

Based on SUMMARY and CONTEXT, does the TARGET sentence describe the defendant as a bad parent? Explain your answer. &
``no'' & 
Incorrect answer and without explanation. \\
\hline

\end{tabular}
\caption{Experiments from zero-shot prompting \texttt{FLAN-T5-large}. Real names have been substituted with fictional names in the snippets shown. }
\label{tab:flan-t5-large}
\end{table*} 

\begin{table*}
\centering
\small
\begin{tabular}{|l|p{7cm}|p{2cm}|p{5cm}|}
\hline
   & Quote & Decisions & Lawyers' reasons \\
\hline
\hline
(A) & It is beyond comprehension to me that this woman could do what she did, and I submit to you that if you watched her during the course of this trial she has shown no remorse.  & 
GA = \xmark \newline
Model = \cmark \newline
Lawyers = \cmark & 
Describing that the defendant has not shown any remorse. \\
\hline
(B) & Q. First of all, Mr. [witness], let's talk about the defendant's attitude and emotional state and what was going on when she came back to your house after [the victim] had left with her. She is accusing [the victim] of taking her money. Is she a little bit more agitated at this point? […] She tore your bedroom apart, didn't she? \newline
A. Yes, she did. […] \newline
Q. So, this calm demeanor that [the defendant] usually had was not present? \newline
A. It wasn't there.  & 
GA = \xmark \newline
Model = \cmark  \newline
Lawyers = \cmark  & 
Portraying the defendant as agitated, not calm.\\
\hline
(C) & [The defendant] said that [the victim] looked really mad, really upset. & 
GA = \xmark \newline
Model = \cmark  \newline
Lawyers = \xmark & 
Describing the emotions of the victim, not the defendant. \\
\hline
(D) & Q: What did [the defendant] say to [the victim] when she said these comments? \newline
A: ``I don\'t know what you are doing all that crying for because all it is is fake ass tears. '' &
GA = \xmark \newline
Model = \cmark  \newline
Lawyers = \xmark  & 
Describing the emotions of the victim, not the defendant.
\\
\hline
(E) & Q. Do you recall what was said? \newline
A. [The defendant] was talking about how she had a lot of food at the house that had been brought in by friends and family members and that she should have brought that to the sheriff's office because she was sure we were hungry and would want something to eat. \newline
Q. Okay. Anything else you remember she said? \newline
A. There was really nothing else, I was so taken back by that, that — after being arrested for the capital murder of her daughter, that she was worried about food.  & 
GA = \cmark \newline
Model = \xmark \newline
Lawyers= \cmark & 
Describing the defendant's lack of emotional response to being arrested for capital murder for her daughter. \\
\hline
(F) & Q. Did you feel personally that [the defendant] didn't want to be around you? What made you feel that way? \newline
A. She was distant at the -- at the wake. She didn't talk to us at all. She didn't try and console us at all. We weren't mean to her, but we just felt like there was --- I don't know, some kind of wall or --- I don't know. & 
GA = \cmark \newline
Model = \xmark \newline
Lawyers = \cmark & 
Describing that the defendant's lack of emotional response and support.\\
\hline
\end{tabular}
\caption{Example passages for \textit{Emotions} where machine decisions deviate from the GA annotator's decision. }
\label{tab:expert-qualitative-emotion}
\end{table*}

\begin{table*}
\centering
\small
\begin{tabular}{|l|llll||llll|||llll|}
\hline
 & \multicolumn{4}{|c|}{\# of positive sentences} & \multicolumn{4}{|c|}{Precision of 3 highest-scored passages} & \multicolumn{4}{|c|}{Precision}  \\
 \hline
Trial &  EMOT & SEX & NORM &  MOM & EMOT & SEX & NORM &  MOM & EMOT & SEX & NORM &  MOM  \\
 \hline
A & 78 &  0 &  80 &  2 &  0.67 &    0.00 &    0.67&   0.00 &  0.24 & 0 & 0.36 &0 \\
B &  45 &  17 & 31 & 21 &  1.00 &   0.00 &    0.33 &  0.00 &  0.2 & 0 & 0.13 & - \\
C &  63 & 106 & 117 &  0& 0.67 &  0.67  &   0.67  &   0.00&  0.46 &0.23 & 0.67 & -  \\
D &  20 &   2 &  11 & 20 &  0.00 &    0.00 &  0.00 &  0.00& 0	&0	&0&	0.1 \\
E & 10 & 89 & 150 & 26 & 0.33 &  1.00 &   0.67 &  0.00 &  0.056 &	0.5& 0.333 &	0.182\\
F & 4 & 0 & 2 & 3 &   0.00 & 0.00 &     0.00& 0.00 & 0 &	0	&0	& 0\\
G & 5 & 76 & 144 & 17 & 0.00 &  0.67 &   0.33 &  0.00& 0 & 0.222 &0.273 & 0 \\
H & 30 &  49 & 258 & 59 &  0.67 &  0.33 &   1.00 &  0.67 &0.26&0.42&0.05&0.09\\
\hline
\end{tabular}
\caption{Passage-level precision broken down by theme and trial. \# of positive sentences refers to sentences annotated positively by GA annotators.}
\label{tab:precisions}
\end{table*}

\begin{table*}
\centering
\small
\begin{tabular}{|l|p{.4\linewidth}|p{.15\linewidth}|p{.3\linewidth}|}
\hline
   & Quote & Decisions & Lawyers' reasons\\
\hline
\hline

(A) & [The defendant] had a job for a period of time as a greeter at [a supermarket]. This was the end of May, into the beginning of July. So about a little more than a month. So that is where she met [her second husband]. She took up with him, and they were married in December. & 
GA = \xmark \newline
Model = \cmark\newline
Lawyers = \xmark& 
Neutral description of the relationship between the defendant with her second husband, does not sexualize the defendant. \\
\hline
(B) & Q. Do you recall [the defendant] speaking to you about [her husband] in any other negative ways? \newline
A. Mostly, she just complained he didn't make enough money. It was never enough. […] \newline
Q. How often would she complain to you about [her husband] not making enough money?\newline
A. I don't know exactly how often, but more often than not. […] Whenever it just wasn't enough for her, she would complain. & 
GA = \cmark\newline
Model = \xmark\newline
Lawyer = \xmark&
This is describing a greedy woman. It discusses the relationships between the defendant and her husband, but does not sexualize the defendant. \\
\hline
(C) & [The defendant] wanted to move in with her mother, and so she moved into a house causing it to be overcrowded. […] So she's now made -- not only has she made the house overcrowded, she now goes and has another baby, making the house even more crowded. & 
GA = \cmark \newline
Model = \xmark\newline
Lawyer = \cmark& 
The tone that the sentence ``\textit{she now goes and has another baby}'' conveys a judgment.\\
\hline
(D) & 
Q: How long were you having sex with your brother's wife?\newline
A: From about five years.\newline
Q: So you had sex with your brother's wife for five years; how many times?\newline
A: Not so much.\newline
Q: What does, ``not so much'' mean; 2, 20, 100? & 
GA = \xmark \newline
Model = \cmark\newline
Lawyer = \textbf{?}&
Context and case knowledge is needed to know who the person answering the question is and whether the brother's wife refers to the defendant. \\
\hline
(E) &
Similar circumstances as it relates to the [the victim's] murder, they were both truck drivers. [The defendant] had a relationship with both of them. [The defendant] wasn't technically married to [xxx], but their relationship was viewed by most as a common law marriage, because they had been together for such a long time and lived together as a married couple. Also they were both shot in the back.&
GA = \xmark \newline
Model = \cmark\newline
Lawyer = \xmark &
The passage describes the facts of the defendant killing her husband. However, it was unclear what ``both of them'' means and why the defendant's other relationship was mentioned in here. \\
\hline
(F) &
Q: How did you leave things with [the defendant]? [...] Was she angry at you? \newline
A: When I have a girlfriend she angry with me.\newline
Q: During the time that you were in [another state], did you try to date other women or have a different girlfriend? [...] What would happen if you try to have a different girlfriend? \newline
A: She stopped right away. \newline & 
GA = \xmark \newline
Model = \cmark\newline
Lawyer = \textbf{?} &  
Need to know the relationship between witness and defendant. Unclear what "she stopped" means and whether it carries significance.  \\
\hline

\end{tabular}
\caption{Example passages for \textit{Hypersexualization} where machine decisions deviates from the GA annotator's decision. }
\label{tab:expert-qualitative-hypersexualization}
\end{table*}

\begin{table*}
\centering
\small
\begin{tabular}{|l|p{.4\linewidth}|p{.15\linewidth}|p{.3\linewidth}|}
\hline
   & Quote & Decisions & Lawyers' reasons \\
\hline
\hline
(A) & [The defendant], the woman who took every opportunity to line her pockets, a heartless schemer who manipulated and lured men to their peril. She deceived her husband, children, family and friends. She has earned the title premeditated murderer, queen of greed and evil. & 
GA = \xmark\newline
Model = \cmark \newline
Lawyers = \cmark& 
Portraying the defendant as a person who manipulates, lies, is greedy and evil, etc. \\
\hline
(B) & Everything had to do with [the defendant's] greed, according to the prosecution. But there's no evidence of that. [...] There's no evidence about a lavish lifestyle. & 
GA = \xmark\newline
Model = \cmark\newline
Lawyers = \xmark\newline
& 
This passage is the defense lawyer's counter argument against portraying the defendant as greedy. \\
\hline
(C) & Q. So, now you are at the police department, and you question this defendant again about her relationship with [her husband], is that correct? \newline
A. Yes. \newline
Q. Did she tell you how she met [her husband]?\newline
A. I don't recall. \newline
Q. Did she tell you how long they had been married? \newline
A. They were married for three years. &
GA = \xmark\newline
Model = \cmark\newline
Lawyers = \xmark& 
Neutral information gathering about the relationships of the defendant and her husband. \\
\hline
(D) & There's nothing about [the defendant’s] existence or her life that is so mitigating that it overcomes the heinous character of this crime that justifies her being allowed to continue to live. When you make the premeditated, calculated decision to slaughter your own family, to commit genocide against your own tribe, to actually commit these acts of genocide in the middle of a courtroom proceeding, you forfeited your claim to life. & 
GA = \cmark\newline
Model = \xmark\newline
Lawyers = \cmark& 
Portraying the defendant as an evil woman who made calculated decisions to slaughter her own family and tribe. \\
\hline
(E) & Q. Now, was there anything characteristic about [the defendant] that would catch your attention?\newline
A. Whenever she would drive her car, she would always have her window down and she would always be smoking her cigarette. And just the mannerism that she would hold her cigarette to the side. \newline
Q. How did she do that? \newline
A. Just by flicking her cigarette. \newline
Q. Any particular style that you refer to it as? \newline
A. Well, as I said before, the kids told me it's like a Hollywood style. & 
GA = \cmark \newline
Model = \xmark\newline
Lawyers = \textbf{?} & 
Doesn't seem to be related, but could be setting up to portray the defendant as performative. \\
\hline
(F) & Q. Did you see any rings? \newline
A. I saw one or two small ones, yes. \newline
Q. What about earrings, i am pointing to my earrings; do you see them? \newline
A. Yes, i do know earrings, yes. \newline
Q. Did you see any earrings? \newline
A. I didn't see earrings. \newline
Q. What else did you see other than these documents, credit cards, the jewelry, three cell phones, and some purses and coins? & 
GA = \cmark\newline
Model = \xmark\newline
Lawyers = \xmark& 
Do not see the relevance of this conversation --- more context and background knowledge about this case is necessary to know why this was originally coded as related to describing the defendant being ``greedy". \\
\hline
\end{tabular}
\caption{Example passages for \textit{Betrayal of Gender Norms} where machine decisions deviates from the GA annotator's decision. }
\label{tab:expert-qualitative-norms}
\end{table*}

\begin{table*}
\centering
\small
\begin{tabular}{|l|p{.4\linewidth}|p{.15\linewidth}|p{.3\linewidth}|}
\hline
   & Quote & Decisions & Lawyers' reasons \\
\hline
\hline
    (A) &
    This woman had been trying to get [a man] to kill her daughter for at least months. It got to the point that he went to this child's father and said [the defendant] won't leave me alone about killing [her daughter].
    & 
    GA = \xmark\newline
    Model = \cmark\newline
    Lawyers = \textbf{?} &
    Factual statements of the case. 
\\
\hline
    (B) & 
    [The defendant] will pose no danger to children if she's punished to life in prison. [The defendant] wanted to be a mom, and she had four kids. 
    Being a mom, that was her dream, but she was never given the necessary tools to do this right. All four of her kids were taken by CPS the same day [the victim] died because the home was so disgusting and unsafe to live in. &
    GA = \xmark\newline
    Model = \cmark\newline
    Lawyers = \cmark
   &
    Portraying the defendant being an unfit mother as ``the home was so disgusting and unsafe to live in.''
\\
\hline 
    (C) &
    You will hear that these parents battered each other and their children; that they neglected them; they abandoned them.
    You will hear that they were gone for days at a time from their home, leaving [the defendant] from a very early age in charge of her brothers and sisters.
    She was, if you will, a mother to her siblings; but, having no role model for that position, had to figure it out for herself, without any supervision or support.
    You will hear that [the defendant] from the beginning acted as buffer between her parents and her siblings, accepting blame and punishment, accepting the screaming, the beatings and the threats, to protect her younger siblings from their parents. & 
    GA = \xmark\newline
    Model = \cmark\newline
    Lawyers = \xmark
    &  
    Describing the defendant's childhood experiences, that her parents are bad parents. 
\\
\hline
    (D) &
    You heard about the defendant's time in prison for previous drug sale convictions. You heard about her children, her son in prison for years and one of her daughters is in prison for five years. And her other daughter is with her grandmother. And we can only hope that there may be some hope for that daughter. & 
    GA = \cmark\newline
    Model = \xmark\newline
    Lawyers = \cmark
    & 
    Implying the outcomes of the defendant's children are attributable to her being a unfit mother. 
\\
\hline
    (E) &
    [The defendant] and her son show up to [the victim’s] s home. […] Now, [the victim] is [her son’s] uncle. And [her son] is just a young man being commanded by her mother to kill [her uncle]. She orders her son to kill her uncle. & 
    GA = \cmark\newline
    Model = \xmark\newline
    Lawyers = \textbf{?}  
    &  
    Factual statements of the crime. 
\\
\hline
\end{tabular}
\caption{Example passages for \textit{Bad mother} where machine decisions deviates from the GA annotator's decision. }
\label{tab:expert-qualitative-mom}
\end{table*}

\end{document}